\pdfoutput=1

\documentclass[11pt]{article}

\usepackage[final]{acl}
\usepackage{graphicx}

\usepackage{times}
\usepackage{latexsym}

\usepackage[T1]{fontenc}

\usepackage[utf8]{inputenc}

\usepackage{microtype}

\usepackage{inconsolata}

\usepackage[normalem]{ulem}
\usepackage{algorithm}
\usepackage{multirow}
\usepackage{amsmath} 
\usepackage{enumitem}

\usepackage{amsmath}
\usepackage{booktabs}
\usepackage{authblk}

\usepackage{algorithm}
\usepackage{algpseudocode}
\usepackage{algorithm}
\usepackage{algpseudocode}

\usepackage[most]{tcolorbox}

\title{Unsupervised Real-Time Hallucination Detection based on the Internal States of Large Language Models}

\usepackage{authblk}
\author[1]{\textbf{Weihang Su}\thanks{swh22@mails.tsinghua.edu.cn}}
\author[1]{\textbf{Changyue Wang}\thanks{contributed equally}}
\author[1]{\textbf{Qingyao Ai}\thanks{Corresponding Author: aiqy@tsinghua.edu.cn}}
\author[1]{\textbf{Yiran Hu}}
\author[2]{\textbf{Zhijing Wu}}
\author[3]{\textbf{Yujia Zhou}}
\author[1]{\textbf{Yiqun Liu}}

\affil[1]{Department of Computer Science and Technology, Tsinghua University}
\affil[2]{School of Computer Science and Technology, Beijing Institute of Technology}
\affil[3]{School of Information, Renmin University of China}

\begin{document}
\maketitle

\begin{abstract}
Hallucinations in large language models (LLMs) refer to the phenomenon of LLMs producing responses that are coherent yet factually inaccurate. 
This issue undermines the effectiveness of LLMs in practical applications, necessitating research into detecting and mitigating hallucinations of LLMs. 
Previous studies have mainly concentrated on post-processing techniques for hallucination detection, which tend to be computationally intensive and limited in effectiveness due to their separation from the LLM's inference process. 
To overcome these limitations, we introduce MIND, an unsupervised training framework that leverages the internal states of LLMs for real-time hallucination detection without requiring manual annotations. 
Additionally, we present HELM, a new benchmark for evaluating hallucination detection across multiple LLMs, featuring diverse LLM outputs and the internal states of LLMs during their inference process. 
Our experiments demonstrate that MIND outperforms existing state-of-the-art methods in hallucination detection\footnote{We have open-sourced all the code, data, and models in GitHub: https://github.com/oneal2000/MIND/tree/main.}. 

\end{abstract}

\section{Introduction}

In recent years, Large Language Models (LLMs) have demonstrated remarkable performance in a variety of natural language processing (NLP) applications~\cite{brown2020language,chowdhery2022palm,touvron2023llama,scao2022bloom,zhang2022opt}.
However, the widespread adoption of LLMs has highlighted a critical problem, i.e., hallucination. 
Hallucination refers to the cases where LLMs generate responses that are logically coherent but factually incorrect or misleading~\cite{maynez2020faithfulness,zhou2020detecting,liu2021token,ji2023survey}. 
Such flaw hurts the effectiveness and robustness of LLMs in real-world NLP applications, underlining the pressing necessity for research on detecting and mitigating hallucinations in LLMs.

Existing studies on hallucination detection for LLM mainly focus on how to identify possible fact-related errors in LLM’s outputs~\cite{lin2021truthfulqa,li2023halueval,manakul2023selfcheckgpt}. 
For example, WikiBio GPT3~\cite{manakul2023selfcheckgpt}, a well-known benchmark for hallucination detection, hired human annotators to annotate a couple of true and false response. 
Hallucination detection methods are then evaluated based on their accuracy in predicting the truthfulness of the answers. 
Based on this paradigm, considerable hallucination detection methods have been proposed with the goal of taking a piece of text as input and predicting whether there are hallucinations in the inputs~\cite{manakul2023selfcheckgpt,zhang2023enhancing,azaria2023internal}. 
Since they detect hallucinations after LLMs finish the inference process, we refer to these hallucination detection methods as post-processing methods.   

Unfortunately, the post-processing methods widely used in existing studies are suboptimal for the applications of hallucination detection models for LLMs in practice.
First, existing post-processing methods often suffer from extreme computation costs and high latency. 
In order to identify hallucinations in input text without ground truth references (otherwise the task would downgrade to a simple fact verification task), hallucination detection models need to be powerful and knowledgeable on their own. 
SOTA detection methods are often implemented with LLMs (e.g., chatGPT, LLaMA, OPT) directly~\cite{manakul2023selfcheckgpt,azaria2023internal,zhang2023enhancing}, which makes the cost of hallucination detection on par or even larger than the inference process of many LLMs.
Second, post-processing methods are intrinsically limited in model capacities.
As post-processing methods detect hallucinations independently with the inference process of LLMs, they can’t analyze how hallucinations are generated from scratches in each LLM.
Some studies tried to bypass this issue through the construction of proxy models~\cite{azaria2023internal}. 
However, such proxy models must be trained with extensive human annotations, otherwise they cannot capture the unique characteristics of each LLM. 
Such manual annotation data are expensive to collect and not preferable considering the rapid developments of LLM techniques.  

To address these limitations, we propose a novel reference-free, unsupervised training framework \textbf{MIND}, i.e., unsupervised \underline{M}odeling of \underline{IN}ternal states for hallucination \underline{D}etection of Large Language Models. 
We highlight MIND with the following advantages:
(1) Unsupervised. In contrast to previous works, MIND is an unsupervised framework that directly extracts pseudo-training data from Wikipedia. 
It doesn’t require any manual annotation for the training of the hallucination detector. 
(2) Real-time. Compared with the existing post-processing methods, MIND is a real-time hallucination detection framework designed to reduce computational overhead and detection latency. With a simple multi-layer perceptron model built upon the contextual embeddings of each token in LLM’s inference process, MIND can conduct the hallucination detection in a real-time process.
(3) Compatibility. MIND is a lightweight framework that can be incorporated into any existing Transformer-based LLMs.


Further, to facilitate future research and to improve the reproducibility of this paper, we introduce a new benchmark for LLM hallucination detection named \textbf{HELM}: \underline{H}allucination detection \underline{E}valuation for multiple L\underline{LM}s. 
In contrast to previous benchmarks that only provide the text generated based on hand-crafted heuristics~\cite{liu2021token,azaria2023internal} or a single LLM~\cite{manakul2023selfcheckgpt,li2023halueval}, HELM provides not only the texts produced by six different LLMs (together with human-annotated hallucination labels) but also the contextualized embeddings, self-attentions and hidden-layer activations recorded in the inference process of each LLM, which could serve as the snapshots of each LLM’s internal states during their inference process.


In summary, the contributions of our paper are as follows:
\vspace{-3mm}
\begin{itemize}[leftmargin=*]

\item We propose MIND, an unsupervised training framework for real-time hallucination detection based on the internal states of LLM. 
\vspace{-3mm}

\item We introduce HELM, a hallucination detection benchmark featuring text from six LLMs and contains the internal states of each LLM during the text generation process.
\vspace{-3mm}

\item We evaluate MIND and existing hallucintion detection baselines with human annotations. The experimental results show that MIND outperforms existing hallucination detection methods. 

\end{itemize}

\begin{figure*}[t]
\centering
    \includegraphics[width=\textwidth]{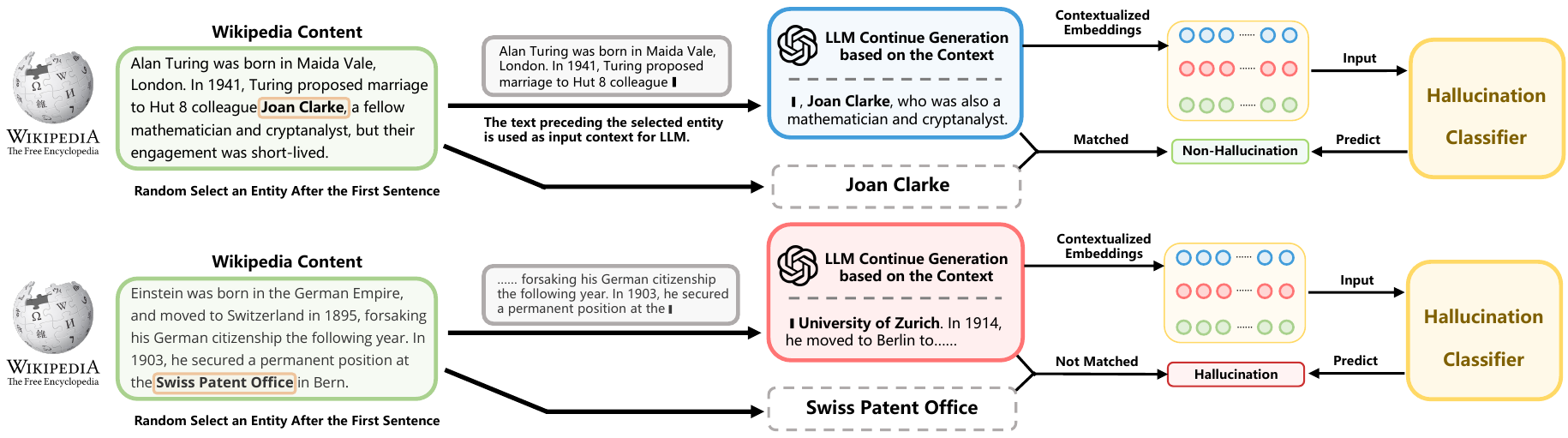}
    \vspace{-4mm}
    \caption{An illustration of the automatic training data generation process of our proposed framework: MIND.}
    \label{pic:framework}
    \label{fig:framework}
\vspace{-4mm}
\end{figure*}
\vspace{-3mm}

\section{Problem Formulation}
\label{sec:problem_detection}

Hallucination Detection can be defined as a binary classification problem. 
The objective is to judge whether the given output from an LLM is a hallucination (false or misleading information) or a non-hallucination (accurate and relevant information). 
This diverges from traditional fact verification tasks. 
In contrast to traditional fact verification tasks that mainly assess the factual accuracy of text, hallucination detection in LLMs goes beyond simply evaluating the truthfulness of the information. 
It focuses more on the comprehensive analysis of the characteristics intrinsic to the generative models. 
Examples of hallucination detection methods include the consistency of multi-responses to the same question~\cite{manakul2023selfcheckgpt}, the LLMs' confidence in its generated content~\cite{zhang2023enhancing}, the internal states of LLMs during the text generation process~\cite{azaria2023internal}, etc.

\section{Methodology}
In this section, we introduce the details of our proposed framework \textbf{MIND}, i.e., unsupervised \underline{M}odeling of \underline{IN}ternal-states for hallucination \underline{D}etection of Large Language Models.
The MIND framework consist of two steps: automatic training data generation and hallucination classifier training.

\subsection{Unsupervised Training Data Generation}
\label{sec:data_generation}
Our proposed Unsupervised Training Data Generation involves automatically annotating hallucinations in content produced by a chosen LLM. 
Let a specific LLM be $L_i$, then our method aims to create customized training data for training a Hallucination Detection Model for $L_i$. 
Figure~\ref{fig:framework} illustrates our proposed automatic data generation process. 
The process starts by selecting a subset of high-quality Wikipedia articles, denoted as $W$ (for example, WikiText-103~\cite{merity2016pointer}). This subset is represented as $W = \{w_1, w_2, ..., w_n\}$, where each $w_i$ ($1 \leq i \leq n$) is an individual article. 
Following that, each article $w_i$ is truncated at a randomly selected entity (except those that appear at the beginning of a sentence) occurring after the first sentence of the article. 
The selected entity is denoted as $e_i$.
The truncated  article is denoted as $w_i^{'} = \text{truncate}(w_i, e_i)$. 
Subsequently, each truncated article $w_i^{'}$ is inputted into the LLM $L_i$. $L_i$ is then tasked with a in free-form text generation based on $w_i^{'}$.
The continuation text produced by the LLM for $w_i^{'}$ is marked as $G_i$ which is truncated at the end of the first sentence.
During the generation of $G_i$, the internal states of the LLM are recorded, denoted as $S_i$. 
Each \( G_i \) is then compared to the original article \( w_i \) to assess if the LLM can accurately continue writing about the original entity \( e_i \). 
The key criterion for this comparison is whether the beginning of \( G_i \) contains the correct entity \( e_i \) as it appears in the original article. 
If \( G_i \) starts with the correct entity \( e_i \), it is labeled as non-hallucination and represented as \( H_i = 0 \). 
Conversely, if \( G_i \) does not start with the correct entity \( e_i \), and \( e_i \) does not appear in \( G_i \), it is labeled as hallucination and represented as \( H_i = 1 \).
For each article, a data tuple is created, represented as $D_i = (L_i, w_i, G_i, S_i, H_i)$, encompassing the selected LLM $L_i$, the original Wiki article $w_i$, generated text $G_i$, the internal states $S_i$, and the hallucination label $H_i$.

\subsection{Hallucination Classifier Training}
\label{sec:training}

\subsubsection{Feature Selection}

In Transformer-based models, the generation of each token is directly based on its contextualized embedding vectors (also named hidden states), which represents the LLM's comprehension and semantic representation of the previously generated tokens. 
These contextualized embeddings also manifest the behavior of the LLM in making next token predictions, encompassing elements of uncertainty. 
Thus, contextualized embedding can serve as an important reference for judging LLMs' Hallucinations.
In this section, we investigate a simple research question:
Can we detect hallucinations in LLMs using contextualized embeddings?

To verify this, we select the contextualized embeddings of different tokens in various Transformer layers and use a multilayer perceptron (MLP) to classify the embedding during hallucinations from those during non-hallucinations. 
If it is possible to classify these vectors using a simple MLP, it indicates distinctions between the token embeddings of LLMs during hallucination and non-hallucination states.
We train this MLP on 5k samples generated by LLaMA2-13B-Chat via the automatic data generation process described in the previous section, and test the prediction accuracy on 5k samples.
The classification accuracy based on the contextualized embedding vectors at various positions of LLM and their combinations are as follows:

\begin{table}[h!]
\setlength\tabcolsep{2.5pt} 
\centering
\footnotesize{\begin{tabular}{lccc}
\toprule
\textbf{Layer}                  & \textbf{Embedding} & \textbf{Formula} & \textbf{Acc} \\
\midrule
\textbf{All}           & All                &     \tiny{$\frac{1}{N}\sum_{j=1}^{N}\frac{1}{n} \sum_{i=1}^{n} {H}_{j}^{i}$ }            & 0.7054       \\
\textbf{First \& Last} & All                &     \tiny{$\frac{1}{2} (\sum_{i=1}^{n} {H}_{1}^{i} + \sum_{i=1}^{n} {H}_{N}^{i})$}             & 0.6929       \\
\textbf{Last}          & All                &     \tiny{$\frac{1}{n} \sum_{i=1}^{n} {H}_{N}^{i}$}          & 0.6986       \\
\textbf{First}         & All                &     \tiny{$\frac{1}{n} \sum_{i=1}^{n} {H}_{1}^{i}$}             & 0.6529       \\
\textbf{Last}          & Last               &     \tiny{${H}_{N}^{n}$}             & 0.7123       \\
\textbf{All}           & Last               &     \tiny{$\frac{1}{N}\sum_{j=1}^{N} {H}_{j}^{n}$}             & 0.6986       \\
\textbf{Last}          & All \& Last        &      \tiny{$\frac{1}{2} (\frac{1}{n} \sum_{i=1}^{n} {H}_{N}^{i} + {H}_{N}^{n}) $ }           & 0.7191       \\
\bottomrule
\end{tabular}}
\end{table}

\noindent In the table above,  $N$ represents the total number of Transformer layers in this LLM, and $n$ is the length of the input token sequence, "${H}_{j}^{i}$" represents the contextualized embedding vector of the $i^{th}$ token in the $j^{th}$ Transformer layer of the LLM.
The experimental results demonstrate that we can indeed classify these vectors using a simple MLP. 
This finding indicates that the contextualized embedding vectors of LLMs exhibit discernible differences during hallucination and non-hallucination states. 
Furthermore, it is noteworthy that an effective distinction can be achieved by merely utilizing the last token's contextualized embedding of the final layer. 
In Section~\ref{sec:experiment}, we extensively explored the generality and robustness of this method in validating existing open-source large models, and conducted more detailed experiments.

\begin{figure}[t]
\centering
    \includegraphics[width=\columnwidth]{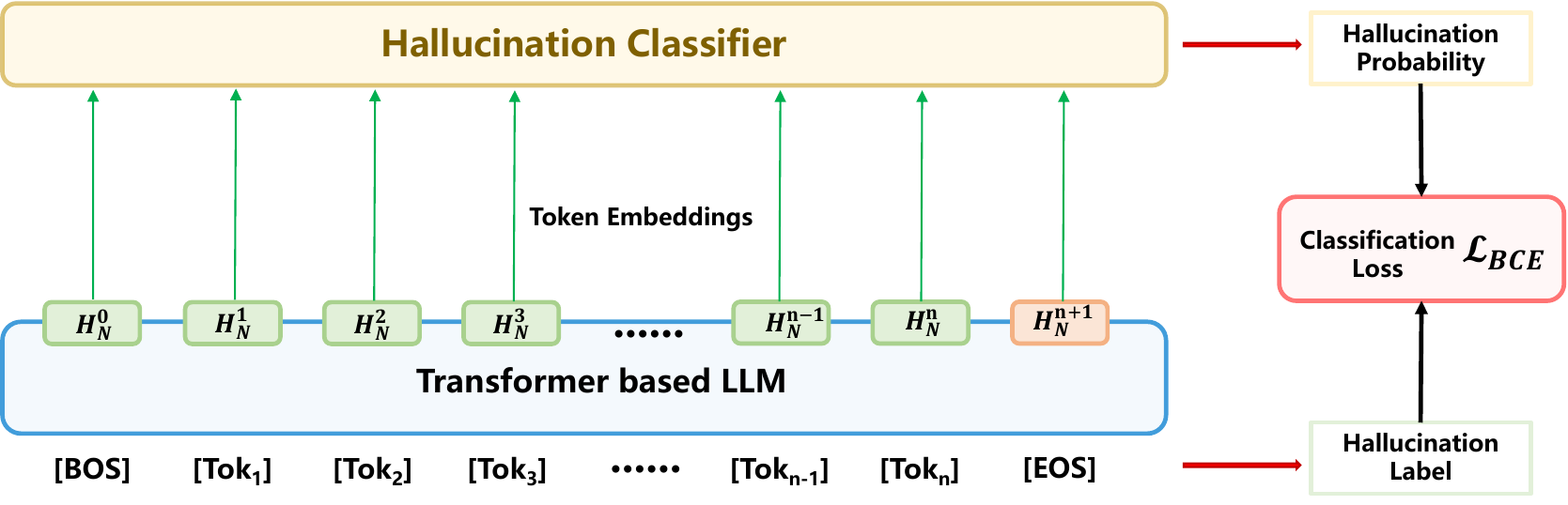}
    \small{\caption{An illustration of the hallucination classifier training process. "${H}_{j}^{i}$" represents the token embedding of the $i^{th}$ token in the $K^{th}$ Transformer layer.}}
    \label{pic:features}
\end{figure}

\subsubsection{Training Process}
The MIND classifier is architecturally structured as a Multilayer Perceptron (MLP) network. 
For the activation functions of the MIND classifier, we select the Rectified Linear Unit (ReLU). 
For the training of the classifier, the input is a \( 1 \times n \) matrix, where \( n \) represents the dimension of the LLM's contextualized embedding. We choose the contextualized embeding of the last token of last Transformer layer as the input of the MIND classifier.
The output of the hallucination classifier is a binary label, indicating whether the LLM is experiencing a hallucination while generating a specific segment of text.
The process can be mathematically represented as follows:
\begin{equation}
\footnotesize
{P} = {MLP}({ReLU}(\mathbf{W} \cdot \mathbf{H} + \mathbf{b})),
\end{equation}
where $P$ is the hallucination label predicted by the classifier, $\mathbf{H}$ is a $1 \times n$ matrix representing the selected hidden states,  $\mathbf{W}$ and $\mathbf{b}$ are the weight matrix and bias vector of the MLP respectively, and MLP represents the function implemented by the multilayer perceptron.
For the loss function, we use the Binary Cross-Entropy (BCE) Loss~\cite{de2005tutorial} to optimize the hallucination classifier, which is defined as:
\begin{equation}
\footnotesize
\mathcal{L}_{BCE}(y_i,p_i) =  y_i \log(p_i) + (1 - y_i) \log(1 - p_i),
\end{equation}
where \( y_i \) represents the actual label of the \( i^{th} \) example, and \( p_i \) indicates the predicted probability that the \( i^{th} \) example belongs to the positive class.

\subsection{Real-time Hallucination Detection}
Upon successfully training the Hallucination Classifier as outlined in Section ~\ref{sec:training}, the MIND framework enable real-time hallucination detection in the chosen LLM. 
Specifically, the Hallucination Classifier receives the contextualized embeddings of each token during the LLM's inference process. 
It then outputs a probability indicating the probability of hallucination in the LLM's output. 

\section{The HELM Benchmark}
\label{sec:helm}

In this section, we introduce the details of our proposed new benchmark HELM: \underline{H}allucination detection \underline{E}valuation for multiple L\underline{LM}s.


\subsection{Data Generation}
We select six widely-used, open-source LLMs for annotation.
These models range in complexity, encompassing both base models and chat models, with sizes varying from 6 billion to 40 billion parameters, including Falcon-40B~\cite{falcon40b}, GPT-J-6B~\cite{gpt-j}, LLaMA-2-Base-7B~\cite{touvron2023llama2}, LLaMA-2-Chat-7B~\cite{touvron2023llama2}, LLaMA-2-Chat-13B~\cite{touvron2023llama2}, and OPT-6.7B~\cite{zhang2022opt}. 
For the generation process, we start with randomly sampling 50,000 articles from a high-quality Wikipedia corpus: WikiText-103~\cite{merity2016pointer}\footnote{The selected articles have no overlap with the training set of MIND.}.
Following that, the selected LLMs were tasked with prompt-based continuation writing. 
For base LLMs and chat LLMs, we have designed different prompt templates respectively which is detailed in Appendix~\ref{append:prompt}.

\subsection{Human Annotation}

Annotators are tasked with evaluating the truthfulness of the content generated by the LLM. 
We only take factual errors into account, without considering grammatical mistakes or subjective opinions\footnote{For example, stating that "Ronald Reagan was a President of the United States" is a fact, whereas commenting on "Ronald Reagan's attractiveness" is a matter of opinion.}.
Furthermore, annotators are instructed not to use generative models (such as ChatGPT) for assistance. 
Wikipedia is the primary source for fact-checking, followed by Google's search engine when Wikipedia lacks relevant information.

The annotation process begins with annotators manually dividing the passage into discrete sentences. 
Annotators are then required to use Wikipedia and Google search to verify the truthfulness of every sentence. 
If neither Wikipedia nor the top 20 search engine results confirm the accuracy of a sentence, it is marked as `Unverifiable.' 
Additionally, in instances of hallucinations, annotators are required to indicate the location of the hallucination within the sentence.
Every sentence and passage in the dataset was annotated by two distinct annotators. 
Only the data with matching annotations from both annotators was adopted. 

\subsection{Benchmark Analysis and Usage}

\begin{table}[]
\caption{ \small{The statistics of the HELM dataset, where H=1 indicates that the sentence is annotated as hallucination, and H=0 indicates non-hallucination. LLB stands for "LLaMA2-Base", and LLC stands for "LLaMA2-Chat".}}
\label{tab:dataset}
\footnotesize{\begin{tabular}{ccccc}
\toprule
                   & \textbf{\#Sentence} & \textbf{\#H=1} & \textbf{\#H=0} & \textbf{\#passage} \\
                   \midrule
\textbf{Falcon-40B} & 521               & 261                    & 260               & 196              \\
\textbf{GPTj-6B}      & 572               & 172                    & 400               & 208              \\
\textbf{LLB-7B}     & 565               & 243                    & 322               & 207              \\
\textbf{LLC-7B}     & 617& 308                    & 309& 204              \\
\textbf{LLC-13B}    & 596& 329                    & 267& 203              \\
\textbf{OPT-7b}     & 566               & 181                    & 385               & 201              \\
\midrule
\textbf{Total}     & 3582& 1494                   & 2088& 1224             \\
\toprule
\end{tabular}}
\end{table}

The statistics of our proposed HELM dataset are shown in Table~\ref{tab:dataset}.
Our dataset ultimately comprises a total of 3342 sentences, each of which corresponds to a label indicating whether it is a hallucination. 
Additionally, it includes the complete contextualized embeddings and hidden layer activations of the LLM during the text generation process.
These 3342 sentences are derived from 1224 distinct passages. 
HELM provides two levels of hallucination detection: Sentence Level and Passage Level.
Since every annotated sentence originates from a passage, {HELM} dataset also encompasses a task for Passage Level Hallucination Detection. The hallucination at the passage level depends on its sentences. If any sentence within a passage is identified as hallucinated, the entire passage is classified as hallucinated.

To use our benchmark, users can either directly use the code provided in our open-source GitHub~\footnote{ We have publicly shared the HELM dataset and code in this anonymous GitHub repository: https://github.com/oneal2000/MIND/tree/main} repository or download our data and implement it themselves.

\section{Experimental Settings}

\subsection{Dataset and Metrics}
We evaluate MIND and other baselines on our proposed dataset HELM which is detailed in section ~\ref{sec:helm}. We use the AUC (Area Under the Curve) and the Pearson correlation coefficient (corr) with human-annotated relevance as evaluation metrics. We conducted experiments at both the sentence-level and the passage-level hallucination detection.


\begin{table*}[]
\caption{The overall experimental results of MIND and other baselines on the HELM benchmark. The best results are in bold. LLB stands for "LLaMA2-Base", SCG stands for "SelfCheckGPT", and LLC stands for "LLaMA2-Chat". Falcon and GPT-J are Falcon-40B and GPT-J-6B, respectively.}
\label{tab:HELM}
\centering
\setlength\tabcolsep{3pt} 
\fontsize{8pt}{8pt}\selectfont
\begin{tabular}{lcccccccccccc}
\toprule
                        & \multicolumn{6}{c}{\textbf{HELM Sentence Level AUC}}                                                      & \multicolumn{6}{c}{\textbf{HELM Passage Level AUC}}                                                       \\
                        \midrule
                        \textbf{Baselines}& \textbf{Falcon} & \textbf{GPT-J} & \textbf{LLB-7B} & \textbf{LLC-13B} & \textbf{LLC-7B} & \multicolumn{1}{l|}{\textbf{OPT-7B}} & \textbf{Falcon} & \textbf{GPT-J} & \textbf{LLB-7B} & \textbf{LLC-13B} & \textbf{LLC-7B} & \textbf{OPT-7B} \\
                        \toprule
\textbf{PE-max}         & 0.6479& 0.7497          & 0.6851           & 0.4439            & 0.4931           & \multicolumn{1}{l|}{0.7263 }          & 0.8347& 0.8875          & 0.8400           & 0.5933            & 0.6988           & 0.8851           \\
\textbf{PE-min}         & 0.5757& 0.7044          & 0.5878           & 0.3164            & 0.4411           & \multicolumn{1}{l|}{0.7228}           & 0.7115& 0.7595          & 0.7587           & 0.5918            & 0.6409           & 0.8075           \\
\textbf{PP-max}         & 0.5749& 0.7074          & 0.5872           & 0.3013            & 0.4166           & \multicolumn{1}{l|}{0.7302}           & 0.7063& 0.7585          & 0.7543           & 0.5344            & 0.6614           & 0.8100           \\
\textbf{PP-min}         & 0.5546& 0.7413          & 0.6025           & 0.3725            & 0.4479           & \multicolumn{1}{l|}{0.7121}           & 0.7649& 0.8526          & 0.7969           & 0.5301            & 0.6870           & 0.8384           \\
\textbf{LNPP}           & 0.5327& 0.6927          & 0.6098           & 0.2673            & 0.3536           & \multicolumn{1}{l|}{0.7206}           & 0.7008& 0.8039          & 0.7755           & 0.5522            & 0.5952           & 0.8374           \\
\textbf{LNPE}           & 0.5442& 0.6980          & 0.6175           & 0.3352            & 0.4114           & \multicolumn{1}{l|}{0.7019}           & 0.7145& 0.8111          & 0.7776           & 0.5878            & 0.6704           & 0.8349           \\
\midrule
\textbf{SCG-MQAG}       & 0.5409& 0.7873          & 0.6401           & 0.4040            & 0.4613           & \multicolumn{1}{l|}{0.7593}           & 0.7275& 0.8827          & 0.8196           & 0.6672            & 0.7284           & 0.8594           \\
\textbf{SCG-NG}    & 0.5218& 0.7549          & 0.5365           & 0.2650            & 0.3105           & \multicolumn{1}{l|}{0.7490}           & 0.7124& 0.8579          & 0.7155           & 0.4983            & 0.5771           & 0.8340           \\
\textbf{SCG-BS}  & 0.6418& 0.7424          & 0.6178           & 0.3026            & 0.3760           & \multicolumn{1}{l|}{0.6594}           & 0.7428& 0.8165          & 0.7631           & 0.4760            & 0.5938           & 0.7597           \\
\textbf{SCG-NLI}        & 0.6846& 0.8680          & 0.7644           & 0.5834            & 0.6565           & \multicolumn{1}{l|}{0.8103}           & 0.8121& 0.9384          & 0.8897           & \textbf{0.7559}   & 0.7951           & 0.9096           \\
\midrule
\textbf{SAPLMA} & 0.5128& 0.6987& 0.5777& 0.3047& 0.4066& \multicolumn{1}{l|}{0.6212}         & 0.7236& 0.8294& 0.7823& 0.5179& 0.6265& 0.7476\\
\textbf{EUBHD} & 0.7509& 0.7593& 0.6479& 0.4658& 0.4805& \multicolumn{1}{l|}{0.7563}         & 0.8659& 0.8560& 0.7859& 0.6685& 0.7126& 0.8545\\
\textbf{GPT4-HDM} & 0.6329& 0.7843& 0.6583& 0.4108& 0.5127& \multicolumn{1}{l|}{0.7972}         & 0.8150& 0.9183& 0.8625& 0.5900& 0.7768& 0.9196\\
\textbf{MIND (ours)}    & \textbf{0.7895}& \textbf{0.8774} & \textbf{0.7876}  & \textbf{0.6043}   & \textbf{0.6755}  & \multicolumn{1}{l|}{\textbf{0.8835}}  & \textbf{0.8886}& \textbf{0.9599} & \textbf{0.9048}  & 0.7175            & \textbf{0.8547}  & \textbf{0.9449} \\
\toprule
\end{tabular}

\begin{tabular}{lcccccccccccc}
                        & \multicolumn{6}{c}{\textbf{HELM Sentence Level Corr}}                                                      & \multicolumn{6}{c}{\textbf{HELM Passage Level Corr}}                                                       \\
                        \midrule
                        \textbf{Baselines}& \textbf{Falcon} & \textbf{GPT-J}  & \textbf{LLB-7B} & \textbf{LLC-13B} & \textbf{LLC-7B} & \multicolumn{1}{l|}{\textbf{OPT-7B}} & \textbf{Falcon} & \textbf{GPT-J}  & \textbf{LLB-7B} & \textbf{LLC-13B} & \textbf{LLC-7B} & \textbf{OPT-7B} \\
                        \toprule
\textbf{PE-max}         & 0.2405& 0.0839         & 0.2032          & 0.2375          & 0.2029           & \multicolumn{1}{l|}{0.0573}          & 0.3106& 0.2660         & 0.2561          & 0.2258           & 0.1956          & 0.2180          \\

\textbf{PE-min}         & 0.1204& -0.0316         & 0.0152          & -0.0166          & 0.0438          & \multicolumn{1}{l|}{0.0601}          & 0.0855& -0.0645         & 0.0928          & 0.2382           & 0.0832          & 0.0823          \\
\textbf{PP-max}         & 0.1193& -0.0467         & 0.0154          & -0.0404          & -0.0337         & \multicolumn{1}{l|}{0.0732}          & 0.0843& -0.0672         & 0.0837          & -0.0426          & 0.1426          & 0.0943          \\
\textbf{PP-min}         & 0.1504& 0.2057          & 0.1842          & 0.1092           & 0.1287          & \multicolumn{1}{l|}{0.1808}          & 0.1815& 0.2454          & 0.2231          & 0.0940           & 0.1668          & 0.1191          \\
\textbf{LNPP}           & 0.0667& 0.0773          & 0.1752          & -0.0971          & -0.0339         & \multicolumn{1}{l|}{0.2075}          & 0.0706& 0.1601          & 0.2297          & 0.2188           & 0.1169          & 0.2417          \\
\textbf{LNPE}           & 0.0936& 0.0376          & 0.1461          & 0.0826           & 0.0858          & \multicolumn{1}{l|}{0.1216}          & 0.1020& 0.1261          & 0.2018          & 0.2867           & 0.2258          & 0.1879          \\
\midrule
\textbf{SCG-MQAG}       & 0.0369& 0.2306          & 0.1822          & 0.1820           & 0.1856          & \multicolumn{1}{l|}{0.2151}          & -0.0851& 0.3145          & 0.2278          & 0.2902           & 0.2581          & 0.2993          \\
\textbf{SCG-NG}    & 0.0995& 0.1770          & 0.0590          & -0.0483          & -0.1016         & \multicolumn{1}{l|}{0.1167}          & 0.0575& 0.2222          & 0.0785          & -0.1065          & -0.1348         & 0.1021          \\
\textbf{SCG-BS}  & 0.2293& 0.0745          & 0.1268          & -0.0136          & -0.0378         & \multicolumn{1}{l|}{-0.0563}         & 0.2471& 0.1288          & 0.1447          & -0.0138          & -0.0504         & -0.0732         \\
\textbf{SCG-NLI}        & 0.3789& 0.4087          & 0.3809          & 0.3092           & 0.3835          & \multicolumn{1}{l|}{0.3312}          & {0.4062}& 0.4217          & 0.4389          & \textbf{0.4145}           & 0.4005          & 0.4091          \\
\midrule
\textbf{SAPLMA} & 0.0208& 0.0456& -0.0015& -0.0003& 0.0130& \multicolumn{1}{l|}{-0.1410}         & 0.0900& 0.1298& 0.0271& -0.0714& 0.0519& -0.1433\\
\textbf{EUBHD} & 0.3161& 0.1057& 0.1170& 0.2115& 0.1043& \multicolumn{1}{l|}{0.1423}         & 0.3447& 0.1546& 0.1693& 0.3079& 0.0512& 0.1342\\
\textbf{GPT4-HDM} & 0.1346& 0.1096& 0.0086& 0.0769& 0.1199& \multicolumn{1}{l|}{0.1678}         & 0.0812& 0.2539& 0.1876& 0.0530& 0.1883& 0.2372\\
\textbf{MIND (ours)}    & \textbf{0.5032}& \textbf{0.5244} & \textbf{0.4857} & \textbf{0.4273}  & \textbf{0.4938} & \multicolumn{1}{l|}{\textbf{0.4760}} & \textbf{0.5251} & \textbf{0.5296} & \textbf{0.4911} & {0.3823}  & \textbf{0.4778} & \textbf{0.5636} \\
\toprule
\end{tabular}
\vspace{-5mm}

\end{table*}

\subsection{Baselines}
We choose the following reference-free hallucination detection methods as baselines:
\begin{itemize}[leftmargin=*]

\item \textbf{Predictive Probability (PP)}~\cite{manakul2023selfcheckgpt}. 
A method for detecting LLMs' hallucinations based on the probability of tokens generated by LLMs.
$PP_{max}$, $PP_{min}$, and LN-PP (Length Normalised PP) indicates max pooling, min pooling and mean pooling methods to combine multiple tokens generated by LLM respectively.


\item \textbf{Predictive Entropy (PE)}~\cite{kadavath2022language} 
\emph{PE} is widely used to evaluate the uncertainty inherent in the model's output distribution~\cite{kadavath2022language}.
For each token \( t_i \) in the LLM output, the \emph{PE} is defined as:
\begin{equation}
\footnotesize
    PE = -\sum_{\tilde{w} \in \mathcal{W}} p_i(\tilde{w}) \log p_i(\tilde{w}) ,
\end{equation}   
\noindent where \( p_i(\tilde{w}) \) represents the likelihood of generating word \( \tilde{w} \), and \( \mathcal{W} \) is the vocabulary of the LLM. $PE_{max}$, $PE_{min}$, LNPE~\cite{malinin2020uncertainty} (Length Normalised PE) indicates max pooling, min pooling and mean pooling strategies to combine multiple tokens generated by LLM respectively.


\item \textbf{SelfCheckGPT}~\cite{manakul2023selfcheckgpt} (SCG) is a strong hallucination detection baseline that is designed based on the principle that if an LLM has knowledge of a given concept, sampled responses are likely to be similar and contain consistent facts. 
SCG employs four distinct techniques: SCG\_BERTScore, SCG\_QA, SCG\_NLI, and SCG\_n-gram, each of which evaluates the consistency of responses from the LLM.





\item \textbf{SAPLMA}~\cite{azaria2023internal} is a novel hallucination detection training method, wherein a classifier is trained using the activation values derived from the hidden layers of LLMs. 

\item \textbf{EUBHD}~\cite{zhang2023enhancing}. {Enhanced Uncertainty-Based Hallucination Detection} is a SOTA hallucination detection method based on the uncertainty of LLMs' outputs.

\item \textbf{GPT4-HDM}~\cite{li2023halueval}. Through a simple prompt template, this method directly use GPT-4 as hallucination detection model to judge whether the text generated by other LLMs contains hallucination. The prompt templates is detailed in Appendix~\ref{append:prompt_GPT4}.

\end{itemize}


\subsection{Implementation Details}
\label{sec:implementation}

\begin{itemize}[leftmargin=*]

\item \textbf{NER}: For the Named Entity Recognition (NER) component of MIND, we follow the methodologies in prior studies~\cite{liu2021token,tarcar2019healthcare}. Specifically, we utilized the Spacy library, a tool recognized for its effectiveness and efficiency in NER as evidenced by previous research~\cite{shelar2020named}.

\item \textbf{MIND}: The MIND classifiers employs a 4-layer Multilayer Perceptron (MLP) network, featuring a 20\% dropout rate applied at the initial layer. The architecture of this network is characterized by a progressively decreasing hidden layer size, with dimensions set at 256, 128, 64, and 2 for each consecutive layer. In terms of activation functions, the Rectified Linear Unit (ReLU) is selected. The learning rate is set to 5e-4, the weight decay is set to 1e-5, and the training batch size is set to 32. For the ablation experiments of important hyperparameters, we have discussed in detail in Section~\ref{sec:ablation}. 


\item \textbf{LLM Configuration}: For the selected LLMs, we directly download model parameters from the official Hugging Face repositories for each model, and use the code provided by Hugging Face to conduct text generation. For the generation configuration, we use the official default configurations provided by each model. We introduce our selected models Appendix ~\ref{sec:selected_llms}.

\end{itemize}

\section{Experimental Results}
\label{sec:experiment}
We conduct experiments to verify our our proposed Hallucination Detection method: MIND. Specifically, this section studies the following research questions (RQ):

\begin{itemize}[leftmargin=*]
\item \textbf{RQ1}: Is hallucination detection based on the internal states of Large Language Models (LLMs) broadly effective across existing LLMs?

\item \textbf{RQ2}: How effective and efficient is MIND on hallucination detection tasks?


\item \textbf{RQ3}: For the training of hallucination detection models, is it necessary to use training data generated by the model itself?


\end{itemize}

\subsection{Overall Results of MIND and Baselines}

In this subsection, we present a comprehensive evaluation of our MIND framework against baselines. The objective is to answer Research Questions RQ1 and RQ2.
Our experimental results are shown in Table ~\ref{tab:HELM}. The key findings are summarized as follows: 
(1) MIND, our proposed unsupervised internal states-based hallucination detection method, demonstrates broad effectiveness in all six LLMs. 
This result validates the hypothesis that hallucination detection based on the contextualized embeddings of LLMs is effective across various models. 
(2) MIND outperforms existing reference-free hallucination detection methods at both sentence and passage levels. This superiority not only highlights the robustness of MIND, but also underscores the effectiveness of the unsupervised training framework proposed in this study. 
(3) SCG-NLI emerged as the second-best hallucination detection method after MIND. Remarkably, it even surpasses MIND in performance on certain models.
However, a critical drawback of SelfCheckGPT is its latency. Specifically, the time taken to detect hallucination in a response is ten times longer than generating the response itself. This makes MIND a more viable option for scenarios requiring real-time hallucination detection. 
(4) SAPLMA, though effective on its own training dataset, exhibits suboptimal performance when applied to the  real-time hallucination detection during the text generation process. This could be attributed to the model's overfitting to the specific training data, which is not generated by the LLM itself.

\subsection{Efficiency}

\begin{table}[]
\caption{Efficiency of MIND and Other Hallucination Detection Baselines. This table presents the average time taken to detect hallucinations in an LLM response and the percentage of time spent on hallucination detection in the complete LLM response time. }
\label{tab:efficiency}
\centering
\footnotesize
\begin{tabular}{lcc}
\toprule
 & \multicolumn{2}{c}{\textbf{LLaMA2-7B-Chat}}\\
\midrule
\textbf{}              & \textbf{Inference Time} & \textbf{Percentage} \\
\toprule
\textbf{LLM's Response}           & 1.52 $s$                       & 100.0 \%                                     \\
\midrule
\textbf{PP \& PE \& EUBHD}       & $<$ 0.01  $s$                     & 0.000      \%                               \\
\textbf{SCG-MQAG}      & 27.06  $s$                     & 1785      \%                           \\
\textbf{SCG-Ngram}   & 15.66   $s$                    & 1033       \%                          \\
\textbf{SCG-BertScore} & 27.29    $s$                   & 1800        \%                         \\
\textbf{SCG-NLI}       & 15.31   $s$                    & 1010         \%                         \\
\textbf{GPT4}       & 4.29   $s$                    & 282.2         \%                         \\
\textbf{MIND }   & 0.05    $s$                    & 3.289            \%                      \\
\toprule
\end{tabular}
\end{table}

This section compares the efficiency of the MIND classifier with other hallucination detection methods, focusing on LLaMA-7B-Chat. 
We examine the average generation time for each question, and the average time of these hallucination detection methods take to detect hallucinations which are demonstrated in Table ~\ref{tab:efficiency}. 
MIND is shown to be highly efficient, taking only 3\% of the LLM's response generation time for hallucination detection. 
This makes MIND ideal for real-time applications, unlike SelfCheckGPT, which is less efficient due to requiring multiple LLM responses. 
Since the LLM records the probability of generating each token during response generation, methods like Predictive Probability (PP) and Predictive Entropy (PE) are easy to compute and require negligible time, making them the least time-consuming. Nonetheless, when considering both efficiency and effectiveness, MIND emerges as the superior reference-free hallucination detection method.

\subsection{Ablation Studies}
\label{sec:ablation}

\subsubsection{Impact of Customized Training Data}
\begin{table}[t]
\caption{Effectiveness of Training OPT-7B and LLaMA2-Chat-7B with Training Data Generated by Different Models. We utilized the HELM dataset and used correlation (Corr) as evaluation metric. The best results are in bold.}
\label{lab:customized}
\setlength\tabcolsep{3pt} 
\scriptsize{
\begin{tabular}{ccccccc}
\toprule
  && \multicolumn{5}{c}{\textbf{LLM Used for Generating Training Data}}\\
                           \midrule
                            & & \textbf{GPTJ} & \textbf{MPT-7B} & \textbf{OPT-7B}  & \textbf{LLB-7B} &\textbf{LLC-7B} \\
                           \toprule
\textbf{Sentence-level}&\textbf{OPT}& 0.4117            & 0.2236          & \textbf{0.4760}  & 0.4136          &0.3312          \\
\textbf{Passage-level}&\textbf{OPT}& 0.5001            & 0.2849          & \textbf{0.5636}  & 0.4981          &0.4349          \\
\midrule
\textbf{Sentence-level}&\textbf{LLC}& 0.2577            & 0.2696          & 0.2357           & 0.3811          &\textbf{0.4938} \\
\textbf{Passage-level}&\textbf{LLC}& 0.3073            & 0.2048          & 0.2159           & 0.3429          &\textbf{0.4778} \\
\toprule
\end{tabular}
}
\end{table}
This section focuses on the importance of incorporating model-specific training data to answer RQ3.
To be specific, we validate the impact of customized training data by comparing the performance of hallucination classifiers in the following two settings. 
In the first setting, training data is customized for the hallucination classifier of each LLM. \textbf{In this setting, the input to the classifier is the internal state of the LLM during its text generation process.}
In the second setting, we use existing training data that was generated by another LLM. Then, this pre-existing data is input into the target LLM. \textbf{In this setting, the input to the classifier is the internal state of the LLM when encoding the pre-existing texts generated by other LLMs.}
Table~\ref{lab:customized} presents the performance of hallucination detection models for OPT-7B and LLaMA-7B-Chat.

The experimental results indicate a clear trend: 
customized training data significantly enhances the performance of hallucination classifiers for both OPT-7B and LLaMA2-Chat-7B models. This observation supports the hypothesis that training data tailored to the specific language model can improve its ability to identify hallucinations in generated text.
This implies that for a new model, it would be advantageous to customize the hallucination detection model using the MIND method tailored to that specific model.

\subsubsection{Impact of the Training Data Size}

\begin{table}[t]
\caption{The performance of MIND at different sizes of training data. We report the accuracy of the dev set.}
\label{tab:scale}
\setlength\tabcolsep{4pt} 
\footnotesize
\centering
\begin{tabular}{cccccc}
\toprule
\textbf{\#Num} & \textbf{1024} & \textbf{2048} & \textbf{3072} & \textbf{4096}   & \textbf{5120} \\
\midrule
\textbf{Accuracy}  & 0.6929        & 0.7066        & 0.7158        & {0.7237} & 0.7192         \\
\toprule
\end{tabular}
\end{table}

The training data for our proposed MIND model is generated automatically, making it crucial to determine the optimal amount of data to produce. We conduct a systematic investigation into the impact of training dataset size on model performance which is detailed in Table ~\ref{tab:scale}. The experimental result shows that increasing the training dataset size from 1,024 to 4,096 labels improves accuracy, suggesting that larger datasets enable the model to learn more complex patterns. However, the benefit plateaus and even slightly decreases after exceeding 4,096 data points, indicating a threshold beyond which additional data no longer improves performance significantly.

\subsubsection{Classifier Layer Depth}

\begin{table}[]
\caption{The performance of MIND at different depths of the classifier. We report the accuracy of the dev set.}
\label{tab:layer}
\setlength\tabcolsep{4pt} 
\footnotesize
\centering
\begin{tabular}{cccccc}
\toprule
\textbf{\#Num} & \textbf{2} & \textbf{3} & \textbf{4} & \textbf{5} & \textbf{6} \\
\midrule
\textbf{Accuracy}    & 0.7157     & 0.7226     & {0.7291}     & {0.7260}      & 0.7248     \\
\toprule
\end{tabular}
\end{table}
In this section, we explore the impact of varying the depth of the classifier layer on the performance of the MIND model. The depth, measured in terms of the number of layers in the classifier, ranged from 2 to 6 layers. The accuracy of the MIND classifier on the development set was used as the metric for assessing performance. The results of this experiment are shown in Table ~\ref{tab:layer}. Initially, as the number of layers increased from 2 to 4, there was a slight improvement in accuracy. Specifically, the model's accuracy went from 0.7157 with two layers to a peak of 0.7291 with four layers. However, the small difference between different numbers of layers indicates that the MIND classifier is not sensitive to this hyperparameter of the number of layers.

\section{Related Work}

\subsection{Hallucination Detection}
To tackle the issue of hallucinations in LLMs, researchers have devised various methods for hallucination detection.  
SelfCheckGPT~\cite{manakul2023selfcheckgpt} (SCG) is designed based on the principle that if an LLM has knowledge of a given concept, sampled responses are likely to be similar and contain consistent facts. 
HaluEval~\cite{li2023halueval} represents a direct approach, where strong LLMs like GPT4 are directly used to evaluate the output of other LLMs. 
SAPLMA~\cite{azaria2023internal} introduces human annotations to label hallucinations ChatGPT outputs, then training a classifier that detects hallucinations in LLMs by analyzing their internal states. 
{EU-HD}~\cite{zhang2023enhancing}. {Enhanced Uncertainty-Based Hallucination Detection} is a SOTA hallucination detection method based on the predictive probability and LLM's attention for each generated tokens.

\subsection{Evaluation of Hallucination Detection}

The purpose of the Hallucination Detection Evaluation (HDE) is to evaluate the effectiveness of various Hallucination Detection Methods (HDMs).
The SelfCheckGPT dataset~\cite{manakul2023selfcheckgpt} utilizes GPT-3 to generate passages about individuals from the WikiBio dataset, with manual annotations assessing the factuality of these passages, classifying them into major inaccurate, minor inaccurate, and accurate categories. 
The True-False dataset~\cite{azaria2023internal} represents another innovative approach, constructed based on a database of instances with multiple factual attributes.  
HaluEval~\cite{li2023halueval} takes a different approach by prompting GPT models to generate hallucinatory texts, using prompts like, “I want you to act as a hallucination answer generator,” coupled with human annotation. 
HADES~\cite{liu2021token} employs a rule-based method to modify tokens in Wikipedia articles to generate hallucination texts.

\section{Conclusions and Future Works}

In this paper, we introduce MIND, a novel unsupervised approach leveraging the internal states of Large Language Models (LLMs) for real-time hallucination detection. 
Moreover, we propose HELM, a comprehensive benchmark for hallucination detection, incorporating outputs from six diverse LLMs along with their internal states during text generation.

\section{Limitations}
We acknowledge the limitations of this paper, particularly in the aspect of detecting hallucinations using only the internal states of LLMs. While effective, this method has the potential for enhanced accuracy. To address this, our future work will focus on integrating the internal states of LLMs with their generated text. This combined approach aims to improve the precision and reliability in identifying and mitigating hallucinations in LLM outputs, leading to more robust and accurate hallucination detection methodologies.

\section{Ethics Statement}
In conducting this research, we have prioritized ethical considerations at every stage to ensure the responsible development and application of AI technologies. 
In the development of MIND, our approach has been to create a reference-free, unsupervised training framework that primarily utilizes publicly available data sources, such as Wikipedia. This methodological choice ensures that our research does not rely on personally identifiable information. 
We firmly believe in the principles of open research and the scientific value of reproducibility. To this end, we have made all models, data, and code associated with our paper publicly available on GitHub. This transparency not only facilitates the verification of our findings by the community but also encourages the application of our methods in other contexts. 

\section*{Acknowledgments}
This work is supported by Quan Cheng Laboratory (Grant No. QCLZD202301).


\bibliography{custom}

\appendix

\section{Pseudocode Description of MIND}
\label{appendix:algorithm}

The pseudocode description of our proposed unsupervised training data generation process is shown in Algorithm ~\ref{algorithm:generation}.

\begin{algorithm*}
\caption{Unsupervised Training Data Generation for Hallucination Detection}
\label{algorithm:generation}
\begin{algorithmic}[1]

\State \textbf{Input:} Language Model $L_i$, Wikipedia Articles $W = \{w_1, w_2, ..., w_n\}$
\State \textbf{Output:} Data Tuples $D = \{D_1, D_2, ..., D_n\}$

\For{each article $w_i \in W$}
    \State Select random entity $e_i$ not at the beginning of any sentence
    \State $w_i^{'} \gets \text{truncate}(w_i, e_i)$ \Comment{Truncate $w_i$ at $e_i$}
    \State $G_i \gets L_i(w_i^{'})$ \Comment{Generate continuation with $L_i$}
    \State Truncate $G_i$ at the end of its first sentence
    \State Record internal states $S_i$ during generation
    \If{the beginning of $G_i$ contains $e_i$ correctly}
        \State $H_i \gets 0$ \Comment{Non-hallucination}
    \Else
        \State $H_i \gets 1$ \Comment{Hallucination}
    \EndIf
    \State $D_i \gets (L_i, w_i, G_i, S_i, H_i)$ \Comment{Form data tuple}
\EndFor

\end{algorithmic}
\end{algorithm*}

\section{Prompt Template of the HELM Dataset}
\label{append:prompt}


For the data generation process of our proposed HELM dataset, the selected LLMs were tasked with free-form generation. 
Specifically, the task involved prompt-based continuation writing. 
For base LLMs and chat LLMs\footnote{Base LLMs are language model that has been pre-trained on a large corpus. On the other hand, the Chat LLM, besides pre-training, also undergoes additional processes such as instruction tuning to better align with conversational tasks.}, we have designed different prompt templates respectively.

For the base LLM, The prompt template is as follows: 
\begin{tcolorbox}[colback=lightgray!20,colframe=darkgray!80,title=Prompt 1]
This is a Wikipedia passage about \textbf{[title]}. \textbf{[First sentence of that article]}.
\end{tcolorbox}
For the chat LLM, The prompt template is as follows:

\begin{tcolorbox}[colback=lightgray!20,colframe=darkgray!80,title=Prompt 2]
The following sentence is the first sentence of a Wikipedia article titled \textbf{[Title]}. Please continue writing the sentence below. \textbf{[First sentence of that article]}
\end{tcolorbox}

\section{Prompt Template of GPT4}
\label{append:prompt_GPT4}

Following the settings of ~\citeauthor{li2023halueval}~\cite{li2023halueval}, we directly use GPT-4 as hallucination detection model to judge whether the text generated by other LLMs contains hallucination through the following prompt template:

\begin{tcolorbox}[colback=lightgray!20,colframe=darkgray!80,title=Prompt 3]
Given the following text span, your objective is to determine if the provided text contains non-factual or hallucinated information. You SHOULD give your judgment based on the world knowledge.

Text span: \textbf{[Provided Text]}

Now, determine if the above text span contains non-factual or hallucinated information. The answer you give MUST be “Yes” or “No”.
\end{tcolorbox}

\section{Details of Our Selected LLMs}
\label{sec:selected_llms}

To validate the effectiveness of the approach utilizing the internal states of Large Language Models (LLMs) for hallucination detection across various existing LLMs, we conducted experiments with as many open-source LLMs as possible. Specifically, this included the following 14 LLMs:

\begin{itemize}[leftmargin=*]


\item \textbf{GPT-J-6B}~\cite{gpt-j} is a 6 billion parameter, autoregressive text generation model trained on The Pile corpus~\cite{pile}.


\item \textbf{OPT-6.7B}~\cite{zhang2022opt}. OPT is a collection of decoder-only pre-trained Transformers, with models ranging from 125 million to 175 billion parameters. We choose \textbf{OPT-6.7B} from the OPT series.






\item \textbf{LLaMA-2}~\cite{touvron2023llama2} is a collection of pre-trained and fine-tuned LLMs ranging in scale from 7 billion to 70 billion parameters. This series includes fine-tuned LLMs, known as Llama 2-Chat, specifically designed for optimal performance in dialogue-based applications. We choose \textbf{LLaMA-2-Chat-7B}, \textbf{LLaMA-2-Base-7B}, \textbf{LLaMA-2-Base-13B}, and \textbf{LLaMA-2-Chat-13}.








\item \textbf{Falcon}~\cite{falcon40b} comprises a set of causal decoder-only models that have been trained on a dataset of 1,000 billion tokens, which includes data from RefinedWeb~\cite{refinedweb}. Among the models in Falcon, we have chosen \textbf{Falcon-7B} and \textbf{Falcon-40B}.

\end{itemize}

\section{Implementation Details}
\label{sec:implementation}

\begin{itemize}[leftmargin=*]

\item \textbf{NER}: For the Named Entity Recognition (NER) component of MIND, we follow the methodologies in prior studies~\cite{liu2021token,tarcar2019healthcare}. Specifically, we utilized the Spacy library, a tool recognized for its effectiveness and efficiency in NER as evidenced by previous research~\cite{shelar2020named}.

\item \textbf{MIND}: The MIND classifiers employs a 4-layer Multilayer Perceptron (MLP) network, featuring a 20\% dropout rate applied at the initial layer. The architecture of this network is characterized by a progressively decreasing hidden layer size, with dimensions set at 256, 128, 64, and 2 for each consecutive layer. In terms of activation functions, the Rectified Linear Unit (ReLU) was selected. The learning rate is set to 5e-4, the weight decay is set to 1e-5, and the training batch size is set to 32. For the ablation experiments of important hyperparameters, we have discussed in detail in Section~\ref{sec:ablation}. 

\item \textbf{SCG}: For the implementation of SelfCheckGPT (SCG), we directly use the code and follow all the hyperparameters from their official GitHub\footnote{https://github.com/potsawee/selfcheckgpt/tree/main}. 
\item \textbf{LLM Configuration}: For the selected LLMs, we directly download model parameters from the official Hugging Face repositories for each model, and use the code provided by Hugging Face to conduct text generation. For the generation configuration, we use the official default configurations provided by each model.

\end{itemize}

\section{Discussion on Mitigating Hallucination}
This paper focuses on hallucination detection and the evaluation of hallucination detection methods. However, in practical applications, it is indeed feasible to use our proposed MIND Framework to mitigate hallucinations. Therefore, we briefly discuss two potential methods to mitigate hallucinations based on the MIND Framework, offering some guidance for those in need.

Firstly, the MIND Classifier can be employed as a re-ranker. The MIND Classifier outputs a score indicating the likelihood of a hallucination occurring in LLMs' response. Thus, by utilizing a sampling decoding method, an LLM can generate multiple outputs for the same query. Subsequently, the MIND Classifier can assess these outputs for hallucinations, allowing the selection of the output with the lowest probability of hallucination as the final response.

Additionally, we can employ the dynamic retrieval augmented generation (RAG) framework~\cite{khandelwal2019generalization,borgeaud2022improving,lewis2020retrieval,guu2020retrieval,izacard2020leveraging,jiang2022retrieval,li2023towards,shi2023replug,su2023caseformer,chen2023thuir,chen2022web,li2023thuir3,su2024dragin,fang2024scaling,zhang2023relevance} to mitigate hallucinations. The dynamic RAG method triggers the retrieval module~\cite{li2023thuir,su2023wikiformer,ma2023caseencoder,ye2024relevance,su2023thuir2} during the inference process of an LLM, facilitating the retrieval of external knowledge. The MIND Framework can provide guidance on when to trigger the RAG. Specifically, when MIND indicates a high probability of hallucination, the retrieval module can be triggered and then add the retrieved passages into the context of the LLM, allowing the LLM to continue generating based on the retrieved external knowledge.

\end{document}